\documentclass[lettersize,journal]{IEEEtran}
\usepackage{amsmath,amsfonts}
\usepackage{algorithmic}
\usepackage{algorithm}
\usepackage{array}
\usepackage{hyperref}
\usepackage[caption=false,font=normalsize,labelfont=sf,textfont=sf]{subfig}
\usepackage{textcomp}
\usepackage{multirow}
\usepackage{tabularray}
\usepackage{stfloats}
\usepackage{url}
\usepackage{verbatim}
\usepackage{graphicx}
\usepackage{cite}
\usepackage{color}

\begin{document}
\title{YOLO-TS: Real-Time Traffic Sign Detection with Enhanced Accuracy Using Optimized Receptive Fields and Anchor-Free Fusion}
% \title{YOLO-TS: Enhancing Real-Time Traffic Sign Detection with Optimized Receptive Fields and Anchor-Free Feature Fusion}

\author{
        Junzhou Chen, Heqiang Huang, Ronghui Zhang, Nengchao Lyu, Yanyong Guo, Hong-Ning Dai, Hong Yan, \textit{Fellow, IEEE}

        \thanks{Our manuscript was first submitted to IEEE Transactions on Intelligent Transportation Systems on June 20, 2024.}
        
        \thanks{This project is jointly supported by National Natural Science Foundation of China (Nos. 61003143, 52172350), Guangdong Basic and Applied Research Foundation (Nos.2021B1515120032, 2022B1515120072), Guangzhou Science and Technology Plan Project (No.2024B01W0079), Nansha Key RD Program(No.2022ZD014), Science and Technology Planning Project of Guangdong Province (No.2023B1212060029). \textit{(Corresponding author: Ronghui Zhang.)}}

        \thanks{
        Junzhou Chen, Heqiang Huang, Ronghui Zhang are with the Guangdong Provincial Key Laboratory of Intelligent Transport System, School of Intelligent Systems Engineering, Sun Yat-sen University, Guangzhou 510275, China. (e-mail: chenjunzhou@mail.sysu.edu.cn; huanghq77@mail2.sysu.edu.cn; zhangrh25@mail.sysu.edu.cn).

        Nengchao Lyu is a professor of Intelligent Transportation Systems Research Center, Wuhan University of Technology, Wuhan 430063, China. (e-mail: lnc@whut.edu.cn).
        
        Yanyong Guo is with the School of Transportation, Southeast University, Nanjing 210097, China. (email: guoyanyong@seu.edu.cn).
        
        Hong-Ning Dai is with the Department of Computer Science, Hong Kong Baptist University, Hong Kong. (E-mail: hndai@ieee.org).

        Hong Yan is with the Department of Electrical Engineering, City University of Hong Kong, Kowloon, Hong Kong. (email: h.yan@cityu.edu.hk).
}
}

% The paper headers
\markboth{IEEE Transactions on Intelligent Transportation Systems, VOL.~XX, NO.~XX,~XXXX}%
{Shell \MakeLowercase{\textit{Chen et al.}}: YOLO-TS: Real-Time Traffic Sign Detection with Enhanced Accuracy Using Optimized Receptive Fields and Anchor-Free Fusion}

\maketitle

\begin{abstract}
Ensuring safety in both autonomous driving and advanced driver-assistance systems (ADAS) depends critically on the efficient deployment of traffic sign recognition technology. While current methods show effectiveness, they often compromise between speed and accuracy. To address this issue, we present a novel real-time and efficient road sign detection network, YOLO-TS. This network significantly improves performance by optimizing the receptive fields of multi-scale feature maps to align more closely with the size distribution of traffic signs in various datasets. Moreover, our innovative feature-fusion strategy, leveraging the flexibility of Anchor-Free methods, allows for multi-scale object detection on a high-resolution feature map abundant in contextual information, achieving remarkable enhancements in both accuracy and speed. To mitigate the adverse effects of the grid pattern caused by dilated convolutions on the detection of smaller objects, we have devised a unique module that not only mitigates this grid effect but also widens the receptive field to encompass an extensive range of spatial contextual information, thus boosting the efficiency of information usage. Evaluation on challenging public datasets, TT100K and CCTSDB2021, demonstrates that YOLO-TS surpasses existing state-of-the-art methods in terms of both accuracy and speed. The code for our method will be available.
\end{abstract}


\section{Introduction}

\begin{figure*}[!t]
    \centering
    \includegraphics[width=180mm]{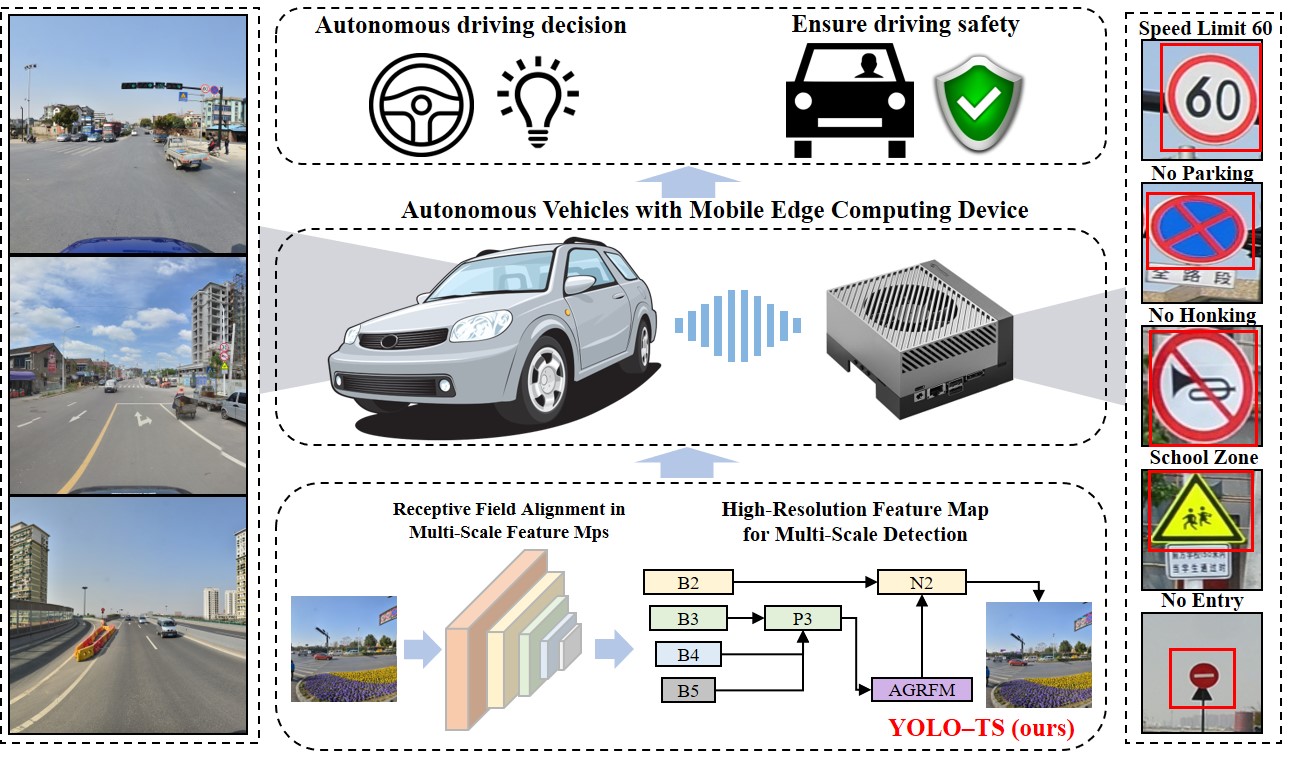}
    \caption{Application Scenarios of Traffic Sign Detection in Autonomous Driving\cite{cleanpng}.}
    \label{Application}
\end{figure*}

% \IEEEPARstart{T}{raffic} signs, crucial components of the transportation system, play a vital role in enabling drivers and autonomous vehicles to accurately capture road information. Their recognition technology provides precise detection results, crucial for avoiding traffic accidents and ensuring road safety. With the advancement of deep learning, notably through sophisticated object detection technologies like Faster R-CNN\cite{ren2016faster} and YOLO\cite{redmon2016you}, there has been a significant enhancement in the performance of object detectors. Particularly, one-stage detectors have been extensively employed across various domains, including traffic sign\cite{yu2022traffic,chen2022real,min2022traffic,wang2023improved}, vehicle\cite{hassaballah2020vehicle, li2022stepwise,charran2022two}, and pedestrian detection\cite{chen2021deep,hsu2023pedestrian,xu2022vehicle,zhou2022sa}, due to their efficient balance between accuracy and speed.
\IEEEPARstart{T}{raffic} signs are crucial components of transportation systems, playing a vital role in enabling drivers and autonomous vehicles to accurately capture road information. As shown in Fig. \ref{Application}, by accurately identifying traffic signs in driving scenarios, autonomous driving systems can make more intelligent and safer driving decisions based on real-time road conditions. This reduces the occurrence of traffic accidents and ensures the safety of both people and vehicles. With the advancement of deep learning, notably through sophisticated object detection technologies like Faster R-CNN\cite{ren2016faster} and YOLO\cite{redmon2016you}, the performance of object detectors has significantly improved. Particularly, one-stage detectors have been extensively employed across various domains, including traffic sign\cite{yu2022traffic,chen2022real,min2022traffic,wang2023improved}, vehicle\cite{hassaballah2020vehicle, li2022stepwise,charran2022two}, and pedestrian detection\cite{chen2021deep,hsu2023pedestrian,xu2022vehicle,zhou2022sa}, due to their efficient balance between accuracy and speed.

% Despite technological advancements, detecting small traffic signs with vehicle-mounted cameras remains a formidable challenge. This difficulty mainly arises from their low resolution and sparse informational content. For example, within an image of $2048 \times 2048$ pixels, a sign may only span an area of $30 \times 30$ pixels. Thus, efficiently detecting such small objects as traffic signs remains one of the substantial challenges in the object detection field.
Despite these technological advancements, detecting small traffic signs with vehicle-mounted cameras remains a formidable challenge. This difficulty primarily arises from their low resolution and limited informational content. For example, within an image of $2048 \times 2048$ pixels, a sign may only span an area of $30 \times 30$ pixels. Thus, efficiently detecting such small objects remains one of the significant challenges in the object detection field.

% To enhance small object detection performance, many studies have focused on creating high-resolution feature maps\cite{cai2021yolov4,zhu2021tph,li2021yolo}. This approach aims to provide more detailed information for accurate predictions. However, these approaches often overlook the importance of aligning the receptive field size with the spatial regions of small objects. Moreover, strategies involving the introduction of a top-down architecture with skip connections to enhance small object detection by integrating low-level details with high-level semantic features across various scales have effectively increased detection accuracy\cite{sun2022rsod,li2023modified,zeng2022small}. Yet, the complexity of these networks has escalated computational costs during training and testing phases, impeding real-time detection capabilities. Therefore, devising a solution that addresses the difficulties of small object detection, computational expense, and the need for real-time processing is urgently required.

To enhance the performance of small object detection, many studies have focused on creating high-resolution feature maps\cite{cai2021yolov4,zhu2021tph,li2021yolo}. This approach aims to provide richer feature representations for accurate predictions. However, these methods often overlook the importance of aligning the receptive field size with the spatial regions of small objects. Additionally, employing top-down architectures with skip connections to merge low-level and high-level features across different scales has improved detection accuracy\cite{sun2022rsod,li2023modified,zeng2022small}. However, the complexity of these networks raises computational costs during both training and testing, hindering real-time detection. Thus, there is a pressing need for a solution that tackles the challenges of small object detection, high computational demands, and real-time processing.

\begin{figure}[!t]
    \centering
    \includegraphics[width=0.5 \textwidth]{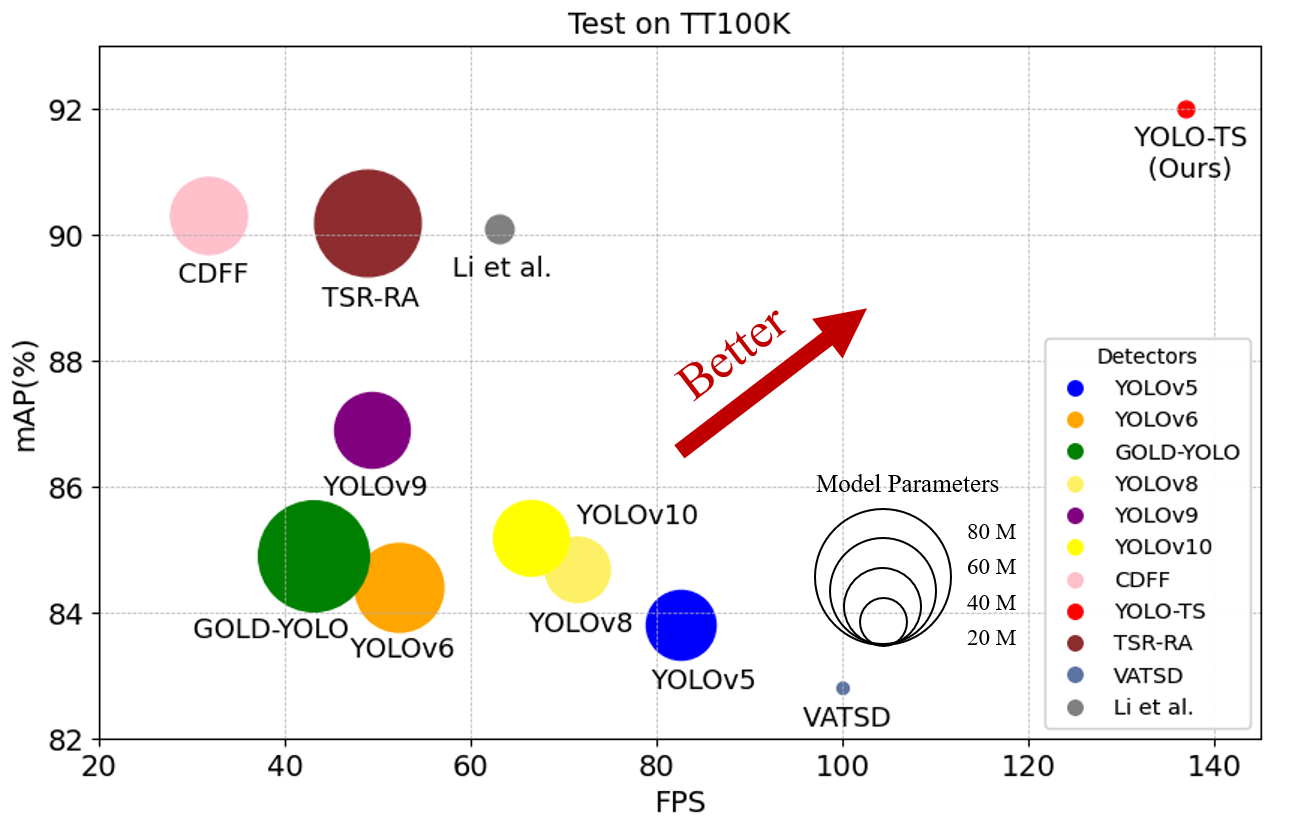}
    \caption{Comparison of the speed and accuracy of different object detectors on TT100K.}
    \label{Comparison}
\end{figure}

To address the challenges of accurately detecting small traffic signs, we present YOLO-TS, a novel small object detection framework inspired by the YOLO (You Only Look Once) series models and specifically optimized for traffic sign detection. This framework aims to significantly enhance both the precision and real-time performance of small object detection, and has demonstrated outstanding performance among various object detectors, as shown in Fig. \ref{Comparison}. By incorporating advanced techniques and optimization strategies, YOLO-TS ensures robust and efficient detection in various driving environments and makes the following contributions to the field of real-time traffic sign detection:

1) \textbf{Receptive Field Alignment in Multi-Scale Feature Maps}: We introduce a novel sensory field-matching strategy that aligns the receptive fields of multi-scale feature maps with the size distribution of traffic signs in the dataset. Addressing the challenge of accurately detecting small objects, this alignment enhances detection precision and speed by ensuring that the receptive fields are optimally configured for various traffic sign sizes.

2) \textbf{High-Resolution Feature Map for Multi-Scale Detection}: To overcome the limitations of traditional multi-scale detection methods, we develop an innovative approach that leverages high-resolution feature maps enriched with contextual information to predict objects across multiple scales. This strategy enhances the flexibility of the anchor-free method, significantly enhancing both precision and speed, and enabling more robust and efficient traffic sign detection in real-time applications.

3) \textbf{Anti-Grid Receptive Field Module (AGRFM)}: In response to the grid effect inherent in dilated convolutions, we design the AGRFM. This module integrates regular convolutions with high dilation rates, enhancing the extraction of small object features by maintaining the continuity of feature maps. Consequently, the overall utilization of spatial information is improved, significantly boosting detection accuracy and reliability.

4) \textbf{Experiment Validation}: Extensive experiments conducted on the TT100K\cite{zhu2016traffic} and CCTSDB2021\cite{zhang2022cctsdb} datasets validate the effectiveness of the proposed YOLO-TS detector. Our method achieves state-of-the-art performance, demonstrating superior accuracy and speed compared to existing approaches. Specifically, YOLO-TS not only surpasses previous methods in mean Average Precision (mAP) but also achieves the highest frames per second (FPS) rate while significantly reducing the model's parameter count.

\section{Related Work}
This section reviews existing work pertinent to our research, with a focus on object detection methodologies, challenges in small object detection, innovations in single-scale feature map prediction, and the use of dilated convolutions.
% In this section, we will review the existing work related to our research, focusing on object detection methodologies, the challenges inherent in small object detection, innovations in single-scale feature map prediction, and the application of dilated convolutions.

\subsection{Object Detection}
In recent years, continuous developments in the field of deep learning have significantly improved the performance of object detection algorithms, leading to the gradual obsolescence of traditional methods. Ross Girshick introduced R-CNN\cite{girshick2014rich} in 2014, marking the first application of CNNs in object detection and pioneering the use of neural networks in this area. Modern object detection algorithms are typically categorized into two types: two-stage detectors, exemplified by the R-CNN series\cite{girshick2014rich,ren2016faster}, and one-stage detectors, exemplified by the YOLO series\cite{jocher2022ultralytics,li2022yolov6,wang2023yolov7,wang2024gold,Jocher_Ultralytics_YOLO_2023,wang2024yolov9,wang2024yolov10}. Two-stage detectors first extract candidate boxes from the image and then classify the contents within these boxes to achieve high-precision object detection. However, this method is relatively slow in detection speed. In contrast, one-stage detectors reformulate the object detection problem as a regression problem, directly predicting the target location and bounding box attributes from image pixels, significantly improving detection speed. This approach is particularly effective in fields such as vehicle, pedestrian, and traffic sign detection. Despite their efficacy, one-stage detectors' proficiency in small object detection remains suboptimal.

\subsection{Small Object Detection}
Detecting small objects is a formidable challenge in object detection because their limited pixel representation makes them distinct from medium and large objects. This pixel limitation often results in a scarcity of feature information, leading to weak feature representations that hinder accurate detection and localization. The small size of these objects further complicates detection, as they can appear anywhere within an image, including in peripheral regions or amidst overlapping objects.

To enhance the detection accuracy of small objects, the research community has explored a variety of strategies. These include data augmentation \cite{zhang2020cascaded,tang2022large,zhou2022small}, multi-scale fusion \cite{qi2022small,hu2022multi,he2022improved,li2022multi}, leveraging contextual information \cite{tang2018pyramidbox,li2022dense,zhang2023attention}, applying super-resolution techniques \cite{rabbi2020small,wang2022remote,zhang2023superyolo,teng2022msr}, and utilizing region proposals \cite{yu2019fruit,cheng2022anchor,lin2020crpn}. Each of these methods aims to empower convolutional neural networks with improved capability for feature extraction from small objects, thereby boosting detection performance.

A critical aspect of enhancing small object detection lies in optimizing the receptive field's size to align with the objects' dimensions and contextual backdrop. An overly small receptive field might fail to encapsulate adequate contextual details, leading to imprecise detection outcomes. Conversely, an excessively large receptive field could inadvertently encompass too much background noise, potentially leading to false positives or imprecise detections. Therefore, fine-tuning the dimensions of the receptive field becomes a crucial strategy for achieving accurate small object detection.

\subsection{Single-Scale Feature Map Prediction}
Traditional YOLO series algorithms\cite{redmon2016you,redmon2018yolov3,bochkovskiy2020yolov4,wang2023yolov7}, typically based on an anchor-based method, generate multiple sets of preset anchor boxes to classify and adjust their positions, covering different sizes and shapes of targets on multi-scale feature maps. Another type of detector relies on the anchor-free method to directly regress the center points and dimensions of targets on different scale feature maps. FSAF\cite{zhu2019feature} dynamically adjusts anchor points on the feature map, significantly enhancing the detection capability for various sizes of targets. YOLOF\cite{chen2021you}, an innovative anchor-free object detection algorithm, introduces a global information guidance module to more effectively utilize global contextual information on a single scale. CenterNet\cite{duan2019centernet} adopts a more streamlined design, combining the target's center point with its width and height, achieving single-scale object detection. Compared to traditional multi-scale methods, the anchor-free method does not depend on predefined anchor boxes but directly predicts the location of targets in the image through the network. Simplifying our model design would be possible if we could directly predict multi-scale targets on a single-scale feature map rich in contextual information.
\subsection{Dilated Convolution}
Dilated convolution is widely used in computer vision tasks such as semantic segmentation and object detection\cite{wang2018understanding}. Its main goal is to expand the receptive field and enhance the capture of contextual information from images while maintaining resolution. In practice, dilated convolution is utilized in two main forms: serial and parallel. Employing mixed dilated convolutions in series, such as Hybrid Dilated Convolution (HDC)\cite{wang2018understanding}, reduces the grid effect through varying dilation rates, while parallel methods like the Atrous Spatial Pyramid Pooling (ASPP) module in DeepLab\cite{chen2017deeplab} and the Receptive Field Block (RFB) module in RFBNet\cite{liu2018receptive} combine convolutions with different dilation rates to fuse multi-scale features. However, these approaches exhibit limitations when addressing small-sized targets, as the grid effect caused by the use of dilated convolutions can lead to discontinuous regions within the feature map. The textural details of small objects may inadvertently fall within these discontinuous regions, adversely affecting the detection of small targets. For small object detection, finding a more effective balance between sensitivity to detail features and maintaining a large receptive field is necessary.

\begin{figure}[!t]
    \centering
    \includegraphics[width=3.2 in]{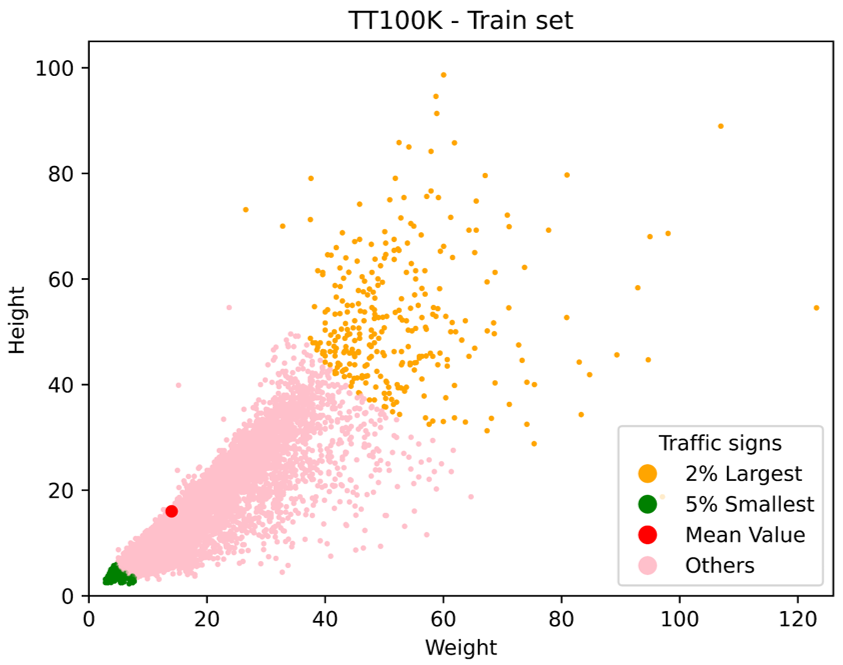}
    \caption{The distribution of object anchor box sizes in the training set of TT100K.}
    \label{anchor box sizes}
\end{figure}

\section{PROPOSED METHOD}
\subsection{Receptive Field Alignment in Multi-Scale Feature Maps}
 
% The efficacy of small object detection hinges on the nuanced alignment of receptive fields with the size distribution of objects. Prior research has extensively explored feature fusion techniques yet often underemphasized the critical alignment between receptive field sizes and small objects in datasets. This alignment is pivotal as objects of varied sizes demand distinct receptive field dimensions for optimal detection\cite{he2019lffd}. 

% High-resolution feature maps, characterized by smaller receptive fields, are inherently adept at detecting smaller objects. Conversely, larger objects are more effectively detected using low-resolution feature maps, which possess larger receptive fields. Through the downsampling process, as feature maps reduce in size, their receptive fields enlarge. However, a mismatch in receptive field size—either too small or too large—can lead to incomplete feature capture or excessive irrelevant information, respectively. Hence, tuning the receptive field sizes at different scales is imperative for effective multi-scale object detection. 

% For this reason, we have precisely calculated the receptive fields of feature maps at different scales to ensure that they effectively match the size distribution of small objects, thereby optimizing the performance of small object detection. The formula for calculating the receptive field of convolutional layers is\cite{cui2020context}:

The efficacy of small object detection hinges on the nuanced alignment of receptive fields with the size distribution of objects. Prior research has extensively explored feature fusion techniques, yet it often underemphasizes the critical alignment between receptive field sizes and small objects in datasets. This alignment is pivotal, as objects of varied sizes demand distinct receptive field dimensions for optimal detection\cite{he2019lffd}.

High-resolution feature maps, characterized by smaller receptive fields, are inherently adept at detecting smaller objects. Conversely, larger objects are more effectively detected using low-resolution feature maps, which possess larger receptive fields. As feature maps are downsampled, their receptive fields enlarge. However, a mismatch in receptive field size—whether too small or too large—can lead to incomplete feature capture or excessive irrelevant information. Hence, tuning the receptive field sizes at different scales is imperative for effective multi-scale object detection.

%%%%%%%%%%%%%%%%%%%%%%%%%%%%%%%
For this reason, we optimally set the receptive fields of feature maps at different scales to ensure they effectively match the size distribution of small objects, thereby optimizing the performance of small object detection. The formula for calculating the receptive field of convolutional layers is as follows\cite{cui2020context}:
\begin{equation}
    \text{RF}_{n} = \text{RF}_{n-1} + (K_{n} - 1) \times \prod_{i=1}^{n-1} S_{i},  
\end{equation}
where $K_{n}$ and $S_{n}$ represent the kernel size and stride of the $n^{\text{th}}$ layer, respectively. $\text{RF}_{n-1}$ is the receptive field size of the previous layer.

Figure \ref{anchor box sizes} illustrates the distribution of anchor box sizes within the TT100K training dataset, guiding the adjustment of receptive field sizes across different network scales. Specifically, the P1 layer is optimized for extracting features from very small signs, while subsequent layers (P2, P3, P4, and P5) are tailored for progressively larger signs. Considering that the practical receptive field tends to be smaller than its theoretical counterpart\cite{luo2016understanding}, we align the receptive fields of multiscale feature maps with the size distribution of smaller targets in the dataset using the following strategy:
\begin{align}
    P1_{\text{RF}} &= \lambda \times \text{anchor}_{\text{tiny}} \\
    P2_{\text{RF}} &= \lambda \times (\text{anchor}_{\text{tiny}} +  \text{anchor}_{\text{mean}})  \\
    P3_{\text{RF}} &= \lambda \times \text{anchor}_{\text{mean}} \\
    P4_{\text{RF}} &= \lambda \times (\text{anchor}_{\text{mean}} +  \text{anchor}_{\text{large}}) \\
    P5_{\text{RF}} &= \lambda \times \ln{(\text{anchor}_{\text{large}})}
\end{align}
where $P1_{\text{RF}}$, $P2_{\text{RF}}$, $P3_{\text{RF}}$, $P4_{\text{RF}}$, and $P5_{\text{RF}}$ represent the theoretical receptive fields of each respective layer. $\lambda$ is a tunable hyperparameter, and $\text{anchor}_{\text{tiny}}$, $\text{anchor}_{\text{large}}$, and $\text{anchor}_{\text{mean}}$ correspond to the average dimensions of the smallest 5\%, largest 2\%, and the overall mean of object sizes in the dataset. 

\begin{table}[]
\centering
\normalsize
\renewcommand{\arraystretch}{1}
\caption{Details of Output Sizes, C2F Blocks, and RF Sizes for YOLO-TS Layers\label{Details}}
\resizebox{0.85\columnwidth}{!}{
\centering
\begin{tabular}{c|ccc}
\hline
Feature map & Output size & C2F blocks & RF size \\ \hline
Input       & 640$\times$640     & -          & 1   \\
P1          & 320$\times$320     & 3          & 27  \\
P2          & 160$\times$160     & 1          & 47   \\
P3          & 80$\times$80       & 1          & 87 \\
P4          & 40$\times$40       & 1          & 167   \\
P5          & 20$\times$20       & 1          & 327  \\ \hline
\end{tabular}
}
\end{table}

% To determine the coefficients in front of \(\lambda\) in the receptive field equations, we performed a detailed statistical analysis of the TT100K dataset. Specifically, we analyzed the size distribution of anchor boxes to identify the optimal receptive field sizes for different feature map layers. The coefficients 0.9, 0.8, 0.7, and 0.6 were empirically chosen based on this analysis to ensure that the receptive fields are appropriately scaled for detecting objects of varying sizes.

% For example, the coefficient 0.9 for \(P2_{\text{RF}}\) was determined by averaging the sizes of the smallest and mean anchor boxes, ensuring that this layer can effectively capture small to medium-sized objects. Similarly, the coefficients for \(P3_{\text{RF}}\), \(P4_{\text{RF}}\), and \(P5_{\text{RF}}\) were chosen to progressively adjust the receptive field size to match the larger objects in the dataset.

% To validate these coefficients, we conducted extensive experiments comparing different sets of coefficients. The results showed that our chosen coefficients significantly improved the mean Average Precision (mAP) for small object detection compared to other combinations. These findings demonstrate that our receptive field alignment strategy effectively balances feature map resolution and receptive field size, optimizing the performance of small object detection.

To validate the choice of $\lambda$, we conducted extensive experiments comparing different values of $\lambda$. The results showed that our chosen $\lambda$ significantly improved the mAP50 for small object detection compared to other values. These findings demonstrate that our receptive field alignment strategy, optimized through the selection of $\lambda$, effectively balances feature map resolution and receptive field size, optimizing the performance of small object detection.

By adjusting the number of Block modules within the C2F module for different scale feature maps in the backbone, we can fine-tune the network depth, thereby altering the size of the receptive fields for feature maps at various scales. Table \ref{Details} provides detailed information on the output sizes for layers P1 to P5, the number of Blocks in the C2F module, the theoretical receptive field sizes, and the detection size ranges. This meticulous adjustment ensures accurate detection across scales, significantly enhancing small object detection precision.

\subsection{High-Resolution Feature Map for Multi-Scale Detection}

\begin{figure*}[!t]
    \centering
    \includegraphics[width=180mm]{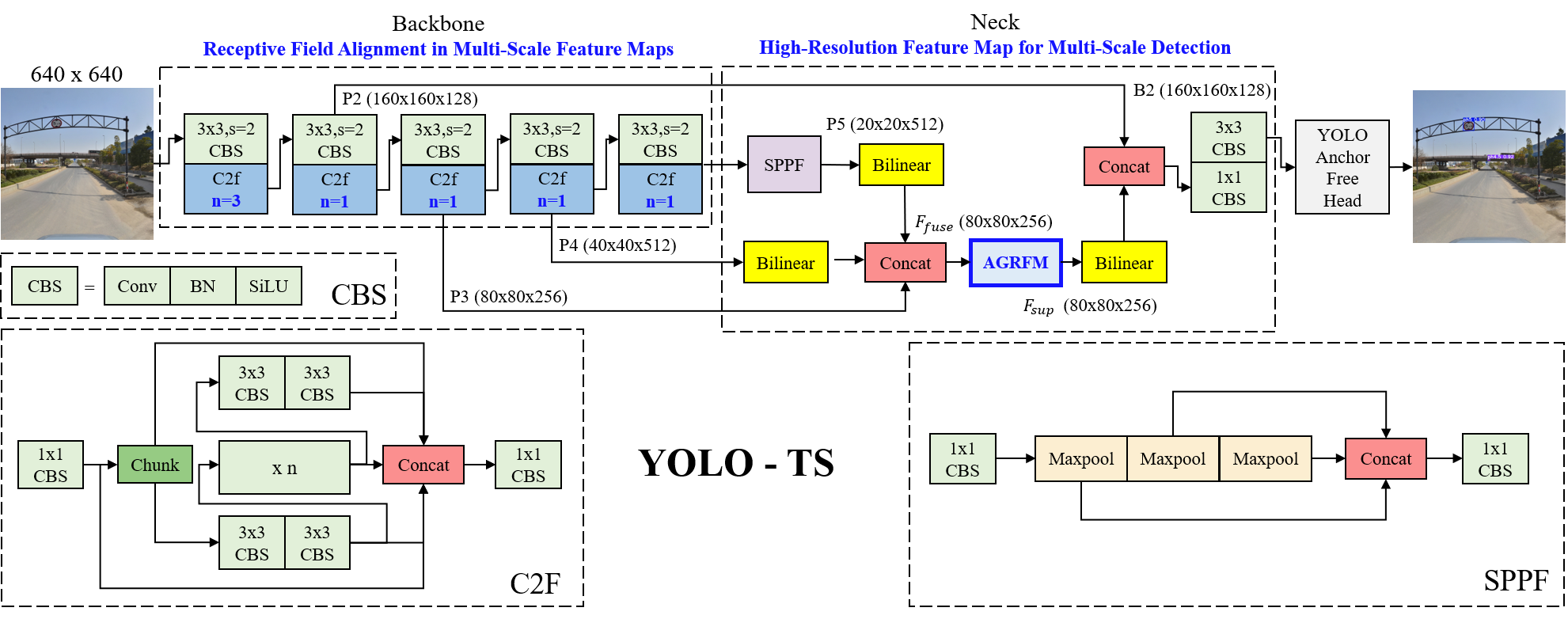}
    \caption{The structure of YOLO-TS.}
    \label{structure}
\end{figure*}

In the realm of convolutional neural networks (CNNs), large-scale feature maps are fundamentally equipped to detect small objects, thanks to their smaller receptive fields which are critical for preserving the granular details of diminutive targets. This attribute makes them inherently preferable for identifying small-scale objects, such as traffic signs, when compared to feature maps derived through successive downsampling, which tend to lose information pertinent to such small entities. However, a drawback of relying solely on large-scale feature maps is their restricted capacity to encapsulate extensive semantic feature information, rendering the direct prediction of multi-scale targets potentially less effective due to the suboptimal extraction of pertinent features. 

Conventionally, a Feature Pyramid Network (FPN)\cite{lin2017feature} is leveraged to enrich the semantic depth of the P2 layer through top-down multi-layer information fusion. However, a limitation of the FPN structure is that it can only fully integrate the features of adjacent layers, and for inter-layer information, it can only be indirectly obtained through a recursive method\cite{wang2024gold}. This causes the information of a certain layer to primarily support its adjacent layers, with limited contributions to other layers, potentially restricting the overall effect of information fusion.

To mitigate the loss of cross-layer information transmission within the FPN, we introduce and enhance an advanced feature fusion mechanism, as depicted in Fig. \ref{structure}. We utilized bilinear interpolation operation to upsample the input features from layers P4 and P5, aligning their size with that of layer P3. By directly merging features from different levels to obtain supplementary layer information, and then infusing this supplementary information into the large-scale feature layer P2 to obtain the final high-resolution feature map B2, which contains rich contextual information. This approach has significantly enhanced the model's detection accuracy for targets of large, medium, and small sizes. The feature fusion mechanism can be summarized as follows:
\begin{align}
    &{F}_{fuse} = \text{Concat}(\text{Bilinear}(P5),\text{Bilinear(P4)},P3)\\
    &{F}_{sup} = \text{AGRFM}({F}_{fuse}) \\
    &{F}_{B2} = \text{Concat}(\text{Bilinear}({F}_{sup},P2))
\end{align}

\subsection{Anti-Grid Receptive Field Module}

Dilated convolution, an extension of traditional convolution, allows networks to expand their receptive fields without increasing computational complexity. However, the use of dilated convolutions with high dilation rates, either singly or in sequence, can lead to a gridding effect. This effect may create regions of discontinuity in feature maps. Some texture details of small objects may fall right into these discontinuous regions, which is detrimental to the detection of small objects.

In contrast, standard convolution layers (non-dilated) can cover continuous pixel areas and capture information between adjacent pixels despite their limited receptive fields. This results in continuous and smooth feature representations on the feature maps. Such smooth representations effectively complement dilated convolutions, providing a seamless transition on the feature map. Therefore, we explore combining the advantages of standard and dilated convolutions to eliminate the gridding effect associated with dilated convolutions.

Employing multiple consecutive standard convolutions effectively acts like a larger convolution kernel through the overlay of smaller kernels. This increases the usage frequency of pixels in the central area compared to those on the edges. Subsequently, applying high dilation rate convolutions spreads the pixel usage frequency. If the dilation rate is too high, the dilated convolution kernels might create multiple, distinctly spaced high-frequency usage areas on the feature map. To prevent the final dilated convolution from creating several small, spaced-out areas, we need to select a moderate dilation rate. This ensures that the final dilated convolution covers no more than all the pixels previously covered by the effective large kernel of the prior standard convolutions. Otherwise, the final dilated convolution will use pixels not covered by the preceding effective large kernel, leading to the phenomenon of multiple closely spaced areas. The following condition should be met:
\begin{equation}
    (k-1) \times r + 1 < k',
\label{equation_10}
\end{equation}
where \(k\) and  \(r\) are the size and the dilation rate of the last dilated convolution kernel. \(k'\) is the equivalent kernel size of all previous standard convolutions.

\begin{figure*}[!t]
    \centering
    \includegraphics[width=7.2 in]{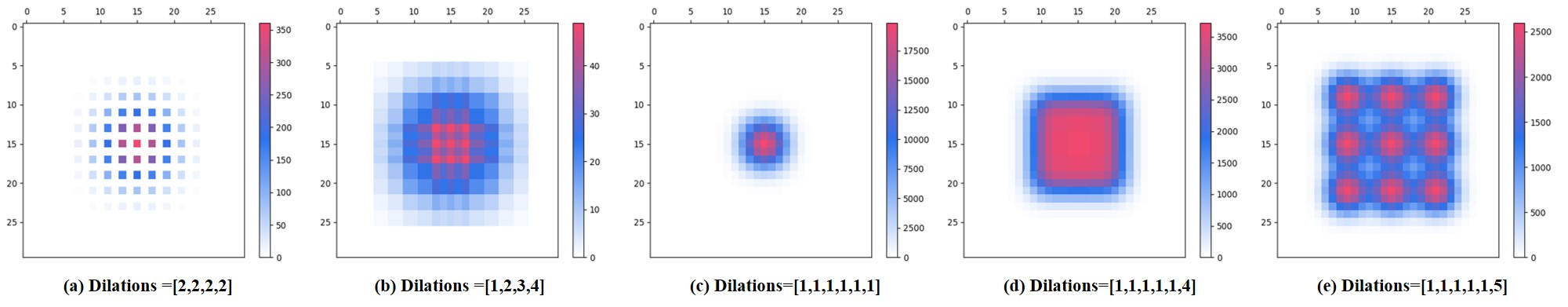}
    \caption{Pixel utilization frequency statistics in feature maps: (a) Pixel utilization frequency statistics in feature map obtained by cascading four $3\times3$ convolutions with a dilation rate of 2. (b) Pixel utilization frequency statistics in feature map obtained by cascading four $3\times3$ convolutions with dilation rates of 1, 2, 3, and 4. (c) Pixel utilization frequency statistics in feature map obtained by cascading six $3\times3$ convolutions with a dilation rate of 1. (d) Pixel utilization frequency statistics in feature map obtained by cascading five $3\times3$ convolutions with a dilation rate of 1 and one $3\times3$ convolution with a dilation rate of 4.(e) Pixel utilization frequency statistics in feature map obtained by cascading five $3\times3$ convolutions with a dilation rate of 1 and one $3\times3$ convolution with a dilation rate of 5.}
    \label{Pixel utilization frequency}
\end{figure*}

Figure \ref{Pixel utilization frequency} illustrates the pixel utilization frequency statistics in feature maps obtained through different combinations of dilated convolution layers. We observe that continuous use of high dilation rates easily produces a significant grid effect. The combination of multiple standard convolutional layers with a single appropriate atrous convolutional layer not only reduces the usage frequency of central pixels but also encourages the frequent utilization of more widely distributed pixels. This effectively avoids the grid effect while enhancing the receptive field while maintaining the continuity of feature maps. If the condition proposed in equation \ref{equation_10} is not met, it will result in the phenomenon of multiple closely spaced areas as shown in Fig. \ref{Pixel utilization frequency}(e).

\begin{figure}[!t]
    \centering
    \includegraphics[width=3.4 in]{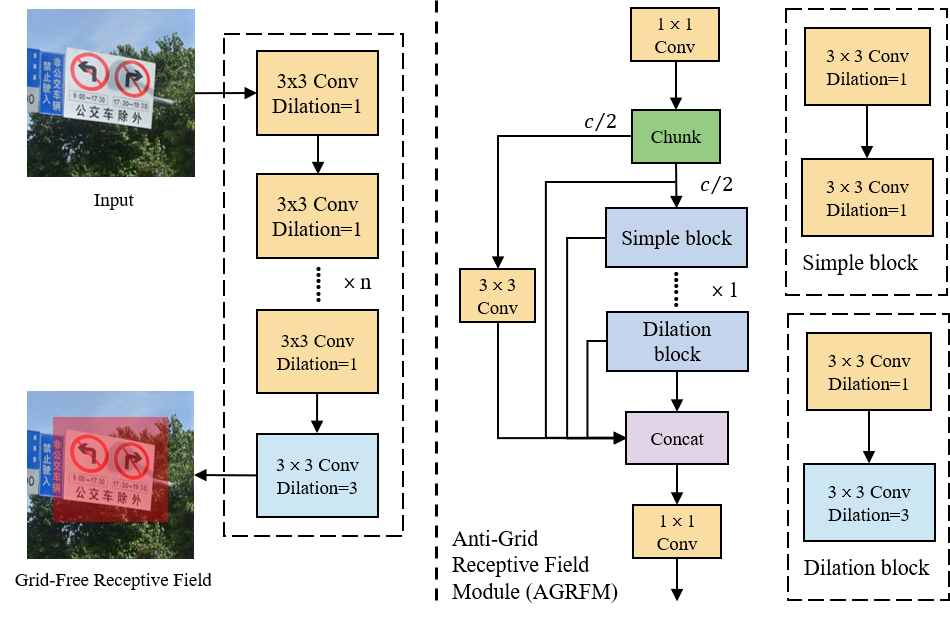}
    \caption{The structure of Anti-Grid Receptive Field Module(AGRFM)}
    \label{AGRFM}
\end{figure}

Inspired by the multi-gradient flow connections in the C2F module, we designed the AGRFM, as shown in Fig. \ref{AGRFM}. We split the input into two paths: one undergoing standard convolution operations and the other passing through a sequence of standard convolution layers followed by a high-dilation-rate convolution layer. This approach mitigates the grid effect introduced by dilated convolutions and achieves information extraction from coarse to fine during the detection process. This strategy not only ensures model efficiency but also enhances the effectiveness of information utilization by capturing a broader context. Consequently, it significantly improves the model's detection capability for targets of various sizes and detail levels.

\section{EXPERIMENTS}

In this section, we conduct a comprehensive validation of our proposed method. We also evaluate the effectiveness of our receptive field alignment in multi-scale feature maps, high-resolution feature map for multi-scale detection, and the AGRFM. We then compare our detector's performance with other state-of-the-art detectors.

\subsection{Experimental settings}

\subsubsection{Datasets}
Our model has been performance-validated across several public benchmark datasets, including TT100K \cite{zhu2016traffic} and CCTSDB2021 \cite{zhang2022cctsdb}. The TT100K dataset, sourced from Tencent Street View Maps, includes 100,000 images with a resolution of 2048×2048 pixels. Of these, 10,000 annotated images feature 30,000 traffic signs. The CCTSDB2021 dataset, developed by Changsha University of Science and Technology in China, comprises 17,856 images in the training and test sets, with traffic signs classified as mandatory, prohibitory, or warning types. There are 16,356 training images, numbered from 00000 to 18991, and 1,500 test images, numbered from 18992 to 20491.

\begin{figure*}[!t]
    \centering
    \includegraphics[width=7.2 in]{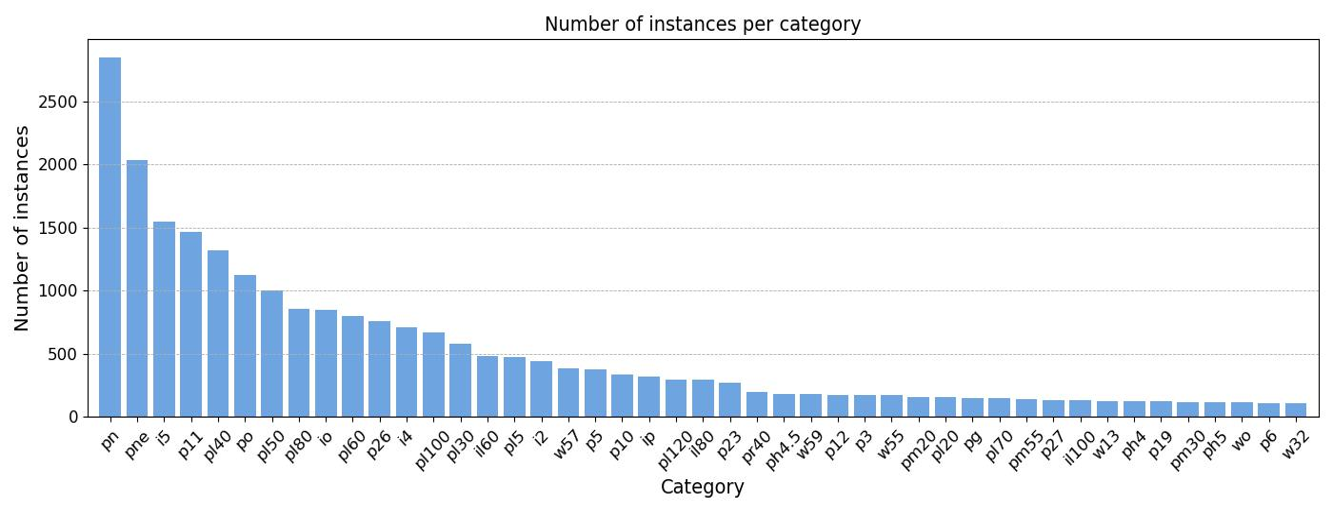}
    \caption{Number of instances per category in TT100K for classes with more than 100 instances}
    \label{Instances}
\end{figure*}

The TT100K dataset comprises around 150 traffic sign categories. Using the approach of Zhu et al. \cite{zhu2016traffic}, we excluded categories with fewer than 100 samples, narrowing the focus to 45 categories. The number of instances for each category is shown in Fig. \ref{Instances}. This benchmark dataset is accessible at \url{http://cg.cs.tsinghua.edu.cn/traffic-sign/}. The training set includes 6,105 images, each with a resolution of 2048×2048 pixels, and approximately 15,000 traffic signs across the 45 categories. The test set consists of 3,071 images with the same resolution, containing 7,070 traffic signs.

\begin{table*}
    \normalsize
    \renewcommand{\arraystretch}{1.3}
    \caption{Performance comparison on TT100K dataset; The first and second best results are indicated in \textbf{\textcolor{blue}{blue}} and \textbf{\textcolor{green}{green}}, respectively.\label{tab:table2}}
    \resizebox{2\columnwidth}{!}{
    \centering
    \begin{tabular}{ccccccccccc}
        \hline
        Method & Venue & Input Size & Precision(\%) & Recall(\%) & F1 & Params(M) & GFLOPs & mAP50(\%) & FPS  & GPU \\ \hline
        {TSR-SA\cite{chen2022real}} & NCA2021 & 608$\times$608  & -  & -  & - & -  & -  & 90.2   & 48.8  & V100 \\
        {CDFF\cite{wang2023cdff}} & NCA2022 & 608$\times$608  & -  & -  & - & -  & -  & \textbf{\textcolor{green}{90.3}}   & 31.8  & TITAN V \\
        {I2D-Net\cite{fu2023small}} & TIM2023 & 512$\times$512  & -  & -  & - & 87.5  & -  & 71.6   & -  & RTX 2070 \\
        {Zhang et al.\cite{zhang2024robust}} & TETCI2024 & 640$\times$640  & -  & -  & - & 81.3  & \textbf{\textcolor{green}{95.0}}  & 70.2   & -  & RTX 2080Ti \\
        {VATSD\cite{wang2023vehicle}} & TITS2024 & 608$\times$608  & -  & -  & - & \textbf{\textcolor{blue}{7.9}}  & \textbf{\textcolor{blue}{16.6}}  & 82.8   & \textbf{\textcolor{green}{100.0}}  & RTX 3080 \\
        {Li et al.\cite{li2024toward}} & TITS2024 & 640$\times$640  & -  & -  & - & -  & -  & 90.1   & 63.0  & TITAN XP \\
        \hline
        {YOLOv5-L\cite{jocher2022ultralytics}} & 2020 & 640$\times$640  & 82.0  & 78.3  & 0.801 & 46.5  & 109.1 & 83.8 & 82.6 & RTX 3090  \\
        {YOLOv6-L\cite{li2022yolov6}} & CVPR2022 & 640$\times$640 & 84.5 & 76.0  & 0.800 & 59.6 & 150.7 & 84.4  & 52.2  & RTX 3090  \\
        {GOLD-YOLO-L\cite{wang2024gold}} & NeurIPS2023 & 640$\times$640  & 83.1 & 77.5 & 0.802 & 75.1 & 151.7 & 84.9 & 43.0 & RTX 3090  \\
        {YOLOv8-L\cite{Jocher_Ultralytics_YOLO_2023}} & 2023 & 640$\times$640 & 83.4  & 77.2  & 0.802 & 43.6  & 165.4  & 84.7  & 71.4 & RTX 4090 \\
        {YOLOv9-C\cite{wang2024yolov9}} & CVPR2024 & 640$\times$640 & \textbf{\textcolor{green}{85.1}} & \textbf{\textcolor{green}{79.0}} & \textbf{\textcolor{green}{0.819}} & 51.1 & 239.4 & 86.5 & 49.3 & RTX 4090 \\
        {YOLOv10-L\cite{wang2024yolov10}} & 2024 & 640$\times$640 & {84.4} & {77.7} & {0.809} & 25.9 & 127.6 & 85.2 & 66.5 & RTX 4090 \\
        \hline
        \textbf{\textcolor{blue}{ours}} & - & 640$\times$640  & \textbf{\textcolor{blue}{89.1$\uparrow$}}  & \textbf{\textcolor{blue}{86.1$\uparrow$}}  & \textbf{\textcolor{blue}{0.876$\uparrow$}} & \textbf{\textcolor{green}{11.1}} & 99.1 & \textbf{\textcolor{blue}{92.0$\uparrow$}} & \textbf{\textcolor{blue}{137.0$\uparrow$}}  & RTX 4090 \\ \hline
    \end{tabular}
    }
\end{table*}

\begin{table*}
    \normalsize
    \renewcommand{\arraystretch}{1.3}
    \caption{Performance comparison on CCTSDB2021 dataset; The first and second best results are indicated in \textbf{\textcolor{blue}{blue}} and \textbf{\textcolor{green}{green}}, respectively.\label{tab:table3}}
    \resizebox{2\columnwidth}{!}{
    \centering
    \begin{tabular}{ccccccccccc}
        \hline
        Method  & Venue  & Input Size & Precision(\%) & Recall(\%) & F1    & Param(M) & GFLOPs & mAP50(\%)     & FPS            & GPU      \\ \hline
        Zhang et al.\cite{zhang2024robust} & TETCI 2024 & 640$\times$640 & - & - & - & 81.3 & \textbf{\textcolor{blue}{95.0}} & \textbf{\textcolor{green}{87.6}} & - & RTX 2080Ti \\
        \hline
        YOLOv5-L\cite{jocher2022ultralytics}  & 2020  & 640$\times$640    & \textbf{\textcolor{blue}{91.3}}          & 75.7       & 0.828 & 46.5     & \textbf{\textcolor{green}{109.1}}  & 82.1          & \textbf{\textcolor{green}{107.5}}          & RTX 4090 \\
        YOLOv6-L\cite{li2022yolov6} & CVPR2022   & 640$\times$640    & 90.0          & 78.6       & \textbf{\textcolor{green}{0.839}} & 59.6     & 150.7  & 84.4          & 75.4           & RTX 4090 \\
        GOLD-YOLO-L\cite{wang2024gold} & NeurIPS2023 & 640$\times$640    & 88.6          & \textbf{\textcolor{green}{79.3}}       & 0.837 & 75.1     & 151.7  & 84.2          & 43.4           & RTX 3090 \\
        YOLOv8-L\cite{Jocher_Ultralytics_YOLO_2023}  & 2023  & 640$\times$640    & 89.0          & 76.6       & 0.823 & \textbf{\textcolor{green}{43.6}}     & 165.4  & 84.3          & 90.9           & RTX 4090 \\
        YOLOv9-C\cite{wang2024yolov9}  & CVPR2024  & 640$\times$640    & 88.0          & 78.9       & 0.832 & 51.1     & 239.4  & 84.6          & 48.3           & RTX 4090 \\
        {YOLOv10-L\cite{wang2024yolov10}} & 2024 & 640$\times$640 & {88.5} & {75.9} & {0.817} & 25.8 & 127.2 & 82.2 & 76.4 & RTX 4090 \\
        \hline
        \textbf{\textcolor{blue}{ours}}    & -   & 640$\times$640    & \textbf{\textcolor{green}{90.1}}          & \textbf{\textcolor{blue}{81.5$\uparrow$}}       & \textbf{\textcolor{blue}{0.856$\uparrow$}} & \textbf{\textcolor{blue}{12.9}}     & 120.2  & \textbf{\textcolor{blue}{88.7$\uparrow$}} & \textbf{\textcolor{blue}{147.1$\uparrow$}} & RTX 4090 \\ \hline
    \end{tabular}
    }
\end{table*}

\subsubsection{Evaluation metrics}
Similar to previous traffic sign detection methods, we evaluate the performance of the proposed algorithm using Precision, Recall, F1 score, mean Average Precision at 50\% IoU (mAP50), and speed (FPS). These metrics are calculated using the following formulas:
\begin{equation}
Precision = \frac{\text{TP}}{\text{TP} + \text{FP}}
\end{equation}

\begin{equation}
Recall = \frac{\text{TP}}{\text{TP} + \text{FN}}
\end{equation}

\begin{equation}
F1 = \frac{2\times {Precision} \times {Recall}}{{Precision} + {Recall}}
\end{equation}

\begin{equation}
\text{mAP50} = \frac{1}{N} \sum_{i=1}^{N} \text{AP}_{50}^{i}
\end{equation}
where TP is the count of true positive traffic sign detections, FP is the count of false positives, FN is the count of false negatives where traffic signs are present but undetected, $N$ represents the total number of categories, and $AP_{50}^{i}$ indicates the average precision at 50\% IoU for each category.

\subsubsection{Training details}
The experimental setup includes a machine equipped with NVIDIA GeForce RTX 4090 GPUs. We use YOLOv8 as our baseline, with methods implemented in PyTorch. The optimizer's learning schedule and settings align with those of YOLOv8, using Stochastic Gradient Descent (SGD) with a 0.01 learning rate and 0.937 momentum. The model is trained for 200 epochs with a batch size of 48, starting from scratch without any pre-trained weights.

\subsection{Comparisons with the state-of-the-arts}
\begin{figure*}[!t]
    \centering
    \includegraphics[width=180mm]{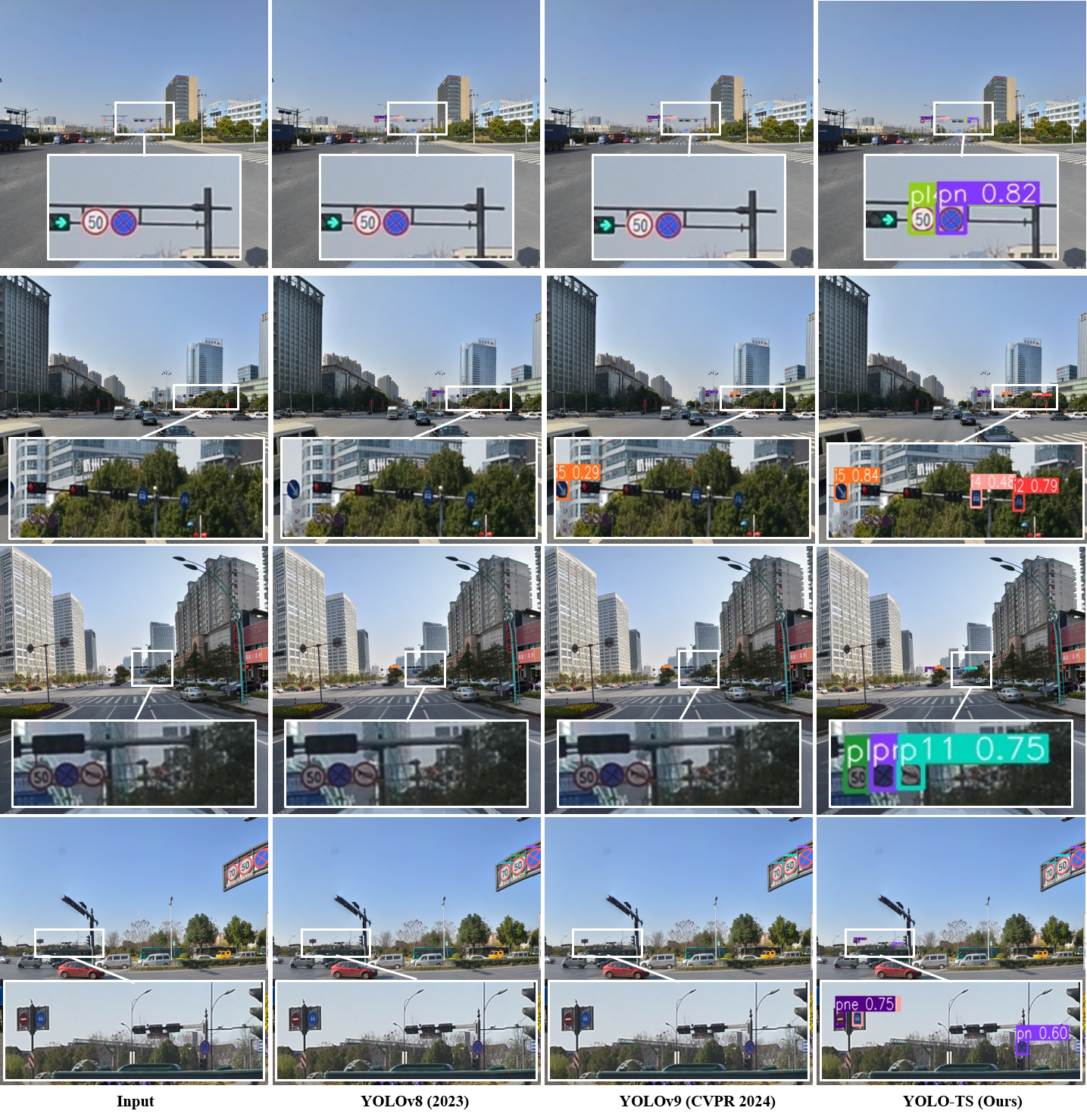}
    \caption{Traffic sign detection results on the TT100K dataset.}
    \label{fig_6}
\end{figure*}

Table \ref{tab:table2} presents the experimental outcomes on the TT100K dataset, demonstrating that our approach attains state-of-the-art performance on multiple critical metrics. Specifically, our model achieved a precision of 89.1\% and a recall of 86.1\%, demonstrating superior recognition accuracy and the ability to reliably detect a greater number of correct traffic sign categories. On the critical mAP50 metric, our approach outperforms the previous best model \cite{wang2023cdff}, which had an mAP50 of 90.3\%, achieving a new high of 92.0\%. This highlights that our algorithm significantly enhances detection accuracy, providing a safeguard for autonomous driving safety. Most notably, compared to the previous state-of-the-art methods, our approach not only significantly improves the mAP50 but also greatly enhances the FPS, achieving an impressive FPS metric of 137.0. Figure \ref{fig_6} presents a comparison of the effectiveness of our method against YOLOv8 and YOLOv9 in detecting traffic signs, which gives us an intuitive impression of the high performance of the proposed method.

Table \ref{tab:table3} shows the experimental results on the CCTSDB2021 dataset. Our model outperforms other advanced YOLO series detectors, including the powerful YOLOv9-C, achieving an mAP50 of 88.7\%. This represents an improvement over the previous best model\cite{zhang2024robust}, which had an mAP50 of 87.6\%. Furthermore, with only 12.9M parameters, our model significantly reduces model complexity while greatly optimizing computational efficiency and resource usage. These results demonstrate the superior performance of our method and its considerable potential for practical applications.

\subsection{Ablation Study}
This section presents ablation experiments to verify the effectiveness of each component in our proposed method, using the challenging TT100K dataset for quantitative analysis.

\begin{table}[]
    \centering
    \normalsize
    \renewcommand{\arraystretch}{1.3}
    \caption{Ablation study for different components of YOLO-TS on TT100K. The best result is marked in \textbf{\textcolor{blue}{blue}}.}
    \label{components}
    \resizebox{1.0\columnwidth}{!}{%
    \begin{tabular}{c|ccc|cccc}
    \hline
    Model  & RFA-M & HR-MSD & AGRFM & Params(M) & GFLOPs & mAP50(\%)& FPS   \\ \hline
    Baseline &  & & & 43.6      & 165.0    & 84.7    & 71.4  \\
             & \checkmark & & & 32.8      & 127.7  & 88.0    & 89.3  \\
            & & \checkmark & & 20.8      & 121.8  & 90.6    & 111.1 \\
          & \checkmark & \checkmark & & 9.8      & 83.8  & 91.5    & 156.3 \\
        & & \checkmark& \checkmark & 21.9      & 137.2  & 90.2    & 101.0  \\
    Ours &\checkmark&\checkmark&\checkmark & 11.1      & 99.1  & \textbf{\textcolor{blue}{92.0$\uparrow$}}    & 137.0 \\ \hline
    \end{tabular}%
    }
\end{table}

\begin{table}[]
    \centering
    \renewcommand{\arraystretch}{1.0}
    \caption{Ablation Study on Different Values of $\lambda$ for Receptive Field Alignment in Multi-Scale Feature Maps. The best result is marked in \textbf{\textcolor{blue}{blue}}.}
    \label{lamda}
    \resizebox{0.8\columnwidth}{!}{%
    \tiny %
    \begin{tabular}{c|ccc}
    \hline
    $\lambda$ & Params(M) & GFLOPs & mAP50(\%)     \\ \hline
    1 & 11.0      & 90.6   & 91.6          \\
    2 & 11.0      & 90.6   & 91.6          \\
    3 & 11.0      & 94.9   & 91.7          \\
    4 & 11.1      & 99.1   & \textbf{\textcolor{blue}{92.0$\uparrow$}} \\
    5 & 11.2      & 103.3  & 91.9          \\
    6 & 12.5      & 111.8  & 91.8          \\ \hline
\end{tabular}%
}
\end{table}

\subsubsection{Effectiveness of Receptive Field Alignment in Multi-Scale Feature Maps (RFA-M)}
To assess the effectiveness of RFA-M, we conducted the corresponding ablation study as shown in the second row of Table \ref{components}. RFA-M maintains a high mAP50 while effectively simplifying the network architecture. Specifically, RFA-M reduces the model's parameters from 43.6M to 32.8M and GFLOPs from 165.0 to 127.7, while increasing the mAP50 to 88.0\% and the FPS to 89.3. This demonstrates that RFA-M not only simplifies the network architecture but also enhances detection performance.

\subsubsection{Different values of $\lambda$ in the Receptive Field Alignment in Multi-Scale Feature Maps (RFA-M)}
To validate the selection of the hyperparameter $\lambda$, we conducted an ablation study. As shown in Table \ref{lamda}, the mAP50 improves as $\lambda$ increases from 1 to 4, peaking at 92.0\% when $\lambda$=4. Beyond this value, the mAP50 decreases despite the number of parameters and GFLOPs increase. This indicates that $\lambda$=4 achieves the optimal balance between model complexity and detection performance. This optimal value ensures that the receptive fields are well-aligned with the size distribution of objects in the dataset, particularly enhancing the detection performance for small objects by capturing sufficient contextual details without introducing excessive background noise.

\subsubsection{Effectiveness of High-Resolution Feature Map for Multi-Scale Detection (HR-MSD)}
To assess the effectiveness of HR-MSD, we conducted corresponding ablation experiments, as indicated in Table \ref{components}. Upon incorporating HR-MSD, both mAP50 and FPS experienced significant improvements, and the FPS impressively reached 111.1. This demonstrates that HR-MSD substantially enhances the model's recognition capability and significantly increases the speed of detection. These results indicate that the use of high-resolution feature maps not only preserves detailed information necessary for accurate detection of small objects but also improves the overall efficiency and responsiveness of the detection model.

\subsubsection{Effectiveness of AGRFM}
To evaluate the effectiveness of AGRFM, we conducted the corresponding ablation experiment as shown in the fourth row of Table \ref{components}. The incorporation of AGRFM significantly expanded the receptive field after multi-scale feature map fusion. As shown in the last two rows of Table \ref{components}, when AGRFM is used in conjunction with the HR-MSD, the mAP50 is improved to 92.0\%, ensuring that the model achieves high-precision detection across a wide range of traffic sign categories. Moreover, this combination also achieves a high FPS of 137.0, demonstrating that the use of AGRFM can maintain a high speed while effectively enhancing accuracy when combined with RFA-M and HR-MSD.

\begin{table}[]
    \centering
    \normalsize
    \renewcommand{\arraystretch}{1}
    \caption{Ablation Study on Different Dilations of the Last Convolution in the Dilation Block. The best result is marked in \textbf{\textcolor{blue}{blue}}.}
    \label{Dilations}
    \resizebox{0.48\textwidth}{!}{%
    \begin{tabular}{c|cccc}
    \hline
    Dilation rates & Precision(\%) & Recall(\%) & F1   & mAP50(\%) \\ \hline
    d=1             & 89.4          & 84.9       & 0.871 & 91.6    \\
    d=2             & 88.5          & 85.9       & 0.872 & 91.8    \\
    d=3             & 89.1          & 86.1       & 0.876 & \textbf{\textcolor{blue}{92.0$\uparrow$}}    \\
    d=4             & 88.7          & 85.5       & 0.871 & 91.7    \\
    d=5             & 87.0          & 86.7       & 0.868 & 91.1    \\ \hline
    \end{tabular}%
    }
\end{table}

\subsubsection{Different dilation rates of the Dilation Block in AGRFM}
To determine the optimal receptive field expansion, we replaced the last dilated convolution in the Dilation Block of AGRFM with dilated convolutions of different dilation rates. The results of varying the dilation rates of the last convolution in the dilation block are presented in Table \ref{Dilations}. Using dilated convolutions with higher dilation rates, the performance of YOLO-TS improved. However, when the dilation rate is too high, the improvement saturates and may even lead to a decrease in accuracy. This is likely due to the fact that a dilation rate of three is sufficient to match the scale of all objects in the images. Furthermore, if the last dilation rate does not meet the criteria of Equation \ref{equation_10}, as shown in Table \ref{Dilations}, when d=5, the detection accuracy actually decreases. This underscores that an excessively high dilation rate can adversely affect the detection accuracy of small objects by causing the grid effect, which negatively impacts the precision of detecting small objects.

\begin{table*}[]
    \normalsize
    \renewcommand{\arraystretch}{1.3}
    \caption{Performance comparison on CCTSDB2021 dataset on the mobile edge device. $*$ means the result after accelerated inference with TensorRT 8 and FP16. The first and second best results are indicated in \textbf{\textcolor{blue}{blue}} and \textbf{\textcolor{green}{green}}, respectively.\label{mobile edge device}}
    \resizebox{2\columnwidth}{!}{
    \centering
\begin{tabular}{ccccccc|c}
\hline
Method      & Venue & Input Size & mAP50(\%) & FPS  & mAP50(\%)*       & FPS*                                  & Device                                  \\ \hline
YOLOv5-L\cite{jocher2022ultralytics}    & 2020                      & 640$\times$640    & 82.1      & \textbf{\textcolor{blue}{41.1}} & 82.0             & \multicolumn{1}{c|}{\textbf{\textcolor{green}{{57.6}}}}             & \multirow{6}{*}{NVIDIA Jetson AGX Orin} \\
YOLOv6-L\cite{li2022yolov6}    & CVPR2022                  & 640$\times$640    & 84.0      & 24.2 & 83.3             & {30.7}             &                                         \\
GOLD-YOLO-L\cite{wang2024gold} & NeurIPS2023               & 640$\times$640    & 83.8      & 20.0 & 83.0             & {28.5}             &                                         \\
YOLOv8-L\cite{Jocher_Ultralytics_YOLO_2023}    & 2023                      & 640$\times$640    & 84.3      & 37.6 & 84.3             & \multicolumn{1}{c|}{48.3}             &                                         \\
YOLOv9-C\cite{wang2024yolov9}    & CVPR2024                  & 640$\times$640    & \textbf{\textcolor{green}{85.2}}      & 17.9 & \textbf{\textcolor{green}{85.2}}             & {44.4}             &                                         \\
YOLOv10-L\cite{wang2024yolov10}    & 2024                  & 640$\times$640    & 82.2      & 27.2 & 82.0             & {\textbf{\textcolor{blue}{80.6}}}      &                                         \\
\cline{1-7}
ours        & -          & 640$\times$640    & \textbf{\textcolor{blue}{88.7$\uparrow$}}      & \textbf{\textcolor{green}{38.8}} & \textbf{\textcolor{blue}{88.6$\uparrow$}}             & \multicolumn{1}{c|}{53.2}             &                                         \\ \hline
\end{tabular}%
}
\end{table*}

\section{Comporision of Inference Speed}
%To evaluate the inference speed of our model, we selected 3016 test images from the TT100K dataset with a batch size set to 1. As shown in Table \ref{tab:table2}, with an input image size of 640x640, our model achieved the fastest inference speed at 105.3 FPS, a 56.8\% increase over the leading YOLOv9 model, which only reached 45.5 FPS under the same conditions.  This substantial increase in processing speed is critically beneficial for the real-time operational demands of Autonomous Driving Systems (ADS) and Advanced Driver-Assistance Systems (ADAS).
\begin{figure}[!t]
    \centering
    \includegraphics[width=3.4 in]{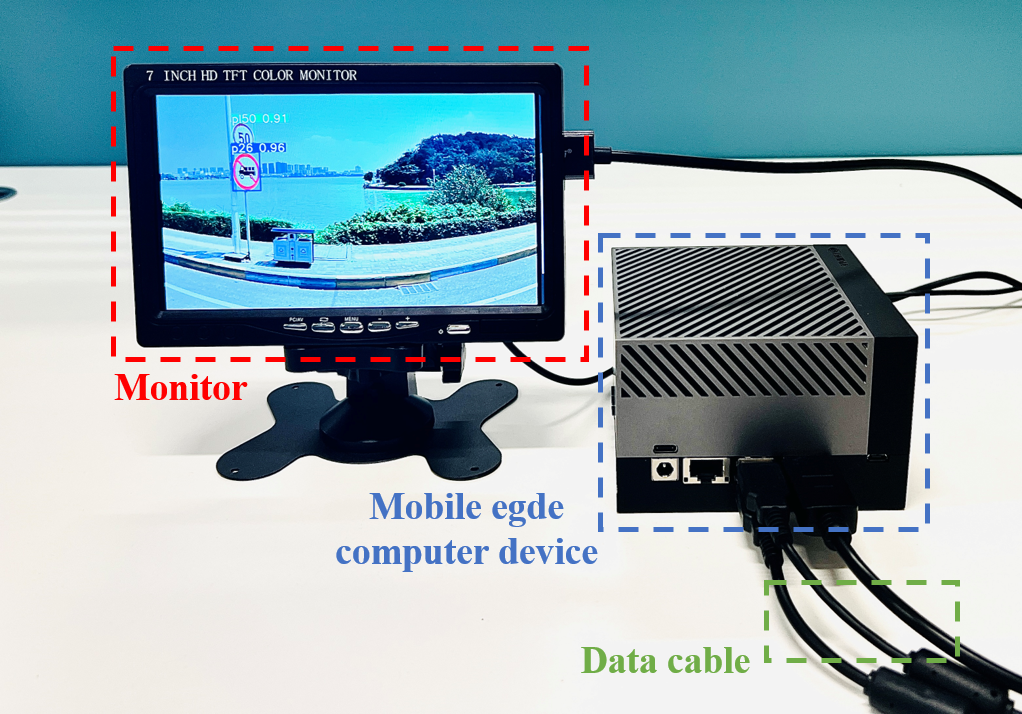}
    \caption{Mobile Edge Computing Device NVIDIA Jetson AGX Orin.}
    \label{egde device}
\end{figure}

In this section, we conducted a comprehensive assessment of the inference speeds of our YOLO-TS model relative to other advanced methods. The tests were conducted on NVIDIA’s high-end GPUs and on the mobile edge device NVIDIA Jetson AGX Orin, as shown in Fig. \ref{egde device}. The latter utilizes the TensorRT 8 FP16 inference acceleration framework. TensorRT can convert PyTorch deep learning models into optimized TensorRT engines, enhancing the inference performance and efficiency of models deployed on NVIDIA edge devices through techniques such as layer fusion, precision calibration, and memory optimization.

Our evaluations utilized a validation set composed of 1,500 images from the CCTSDB-2021 dataset, with batch size consistently set at 1. The results, presented in Table \ref{tab:table3} and Table \ref{mobile edge device}, demonstrate that YOLO-TS not only achieves the highest accuracy but also delivers formidable real-time inference speeds on various platforms. Specifically, on NVIDIA GPUs, YOLO-TS achieves an inference speed of up to 147.1 FPS, outperforming models such as YOLOv9-C and YOLOv10-L. On the NVIDIA Jetson AGX Orin, with the support of TensorRT 8 FP16 optimization, YOLO-TS achieved 53.2 FPS. This substantial boost in processing speed, particularly on edge devices, is vital for real-time applications in Autonomous Driving Systems (ADS) and Advanced Driver-Assistance Systems (ADAS), where fast and precise data processing is imperative.

\section{Conclusion and Future Work}

In this study, we introduced YOLO-TS, an efficient and real-time traffic sign detection network inspired by the YOLO (You Only Look Once) series models, specifically designed for detecting small traffic signs. Our design emphasizes the crucial role of the receptive field in detecting small objects, optimizing the model architecture by aligning the receptive field with the size of small targets to enhance detection speed and accuracy. By fusing multi-scale feature map receptive fields into a single high-resolution feature map rich in contextual information, we leverage the flexibility of the anchor-free approach to support precise detection of multi-scale targets. Additionally, we mitigated the potential grid effect caused by atrous convolutions through the Anti-Grid Receptive Field Module (AGRFM), which combines multiple standard convolution layers with a single dilation convolution layer. This approach enhances pixel utilization and maintains feature map continuity, significantly improving detection accuracy. Our experiments on the TT100K and CCTSDB2021 datasets show that YOLO-TS achieves state-of-the-art performance, significantly enhancing traffic sign detection for real-time applications like autonomous driving and advanced driver-assistance systems.

In future work, expanding the dataset to include more diverse and challenging scenarios will be crucial, such as adverse weather conditions, road signs from different countries, and dynamically changing environments. This expansion will enhance the model's robustness and generalization, ensuring its effectiveness across different regions and in suboptimal conditions. By increasing the diversity of the dataset, we can further optimize and adjust the network to cope with the complex and variable real-world application scenarios, thereby providing more accurate and reliable road sign detection in autonomous driving and advanced driver-assistance systems.

\textbf{We would like to release our source code.}

\ifCLASSOPTIONcaptionsoff
  \newpage
\fi

\noindent
\bibliographystyle{IEEEtran}
\bibliography{Ref.bib}

\vspace{-9 mm}
\begin{IEEEbiography}[{\includegraphics[width=1in,height=1.25in,clip,keepaspectratio]{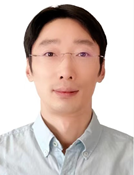}}]
{Junzhou Chen} received his Ph.D. in Computer Science and Engineering from the Chinese University of Hong Kong in 2008, following his M.Eng degree in Software Engineering and B.S. in Computer Science \& Applications from Sichuan University in 2005 and 2002, respectively. Between March 2009 and February 2019, he served as a Lecturer and later as an Associate Professor at the School of Information Science and Technology at Southwest Jiaotong University. He is currently an associate professor at the School of Intelligent Systems Engineering at Sun Yat-sen University. His research interests include computer vision, machine learning,intelligent transportation systems,  mobile computing and medical image processing.
\end{IEEEbiography}

\vspace{-9 mm}
\begin{IEEEbiography}[{\includegraphics[width=1in,height=1.25in,clip,keepaspectratio]{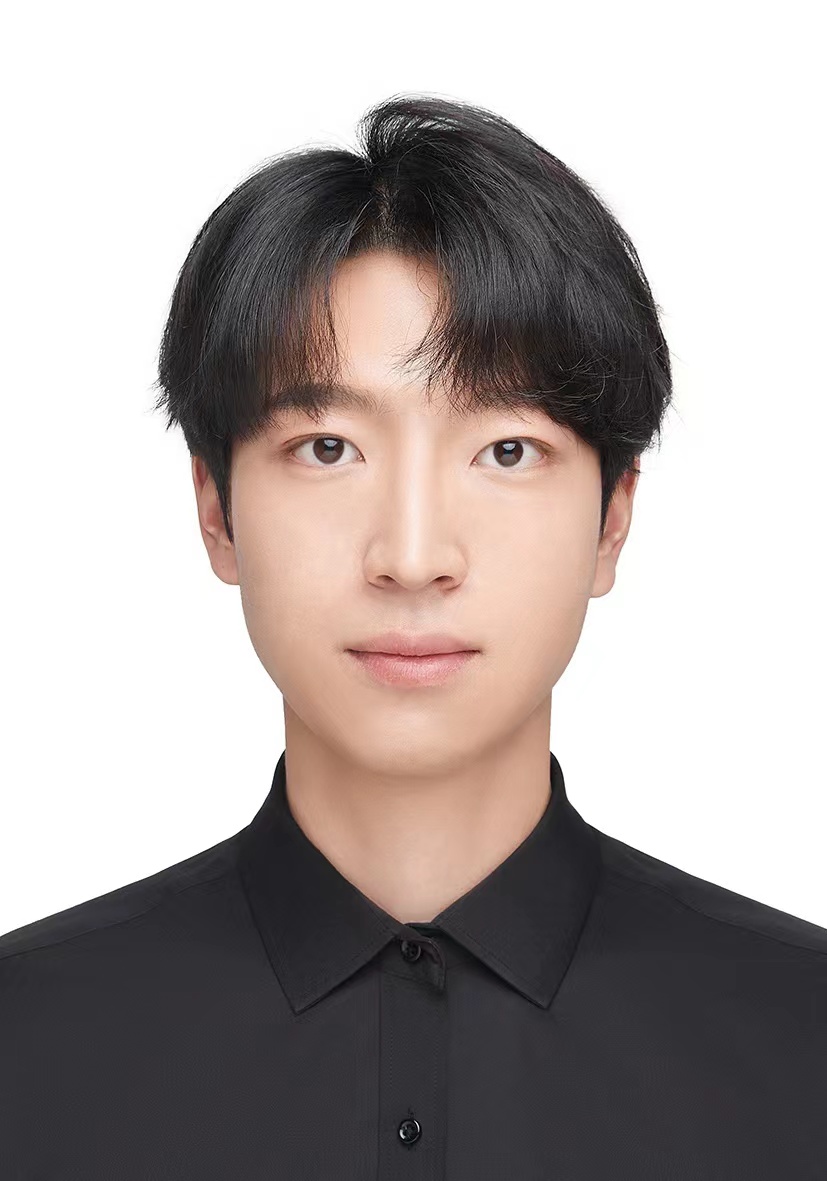}}]
{Heqiang Huang} completed his B.Sc. in Electronic Information Science and Technology from Lanzhou University in 2023. He is presently pursuing his postgraduate studies in Transportation Engineering at the School of Intelligent Engineering, Sun Yat-sen University. Huang's research domain primarily revolves around computer vision, deep learning, and autonomous driving technologies.
\end{IEEEbiography}

\vspace{-9 mm}
\begin{IEEEbiography}[{\includegraphics[width=1in,height=1.25in,clip,keepaspectratio]{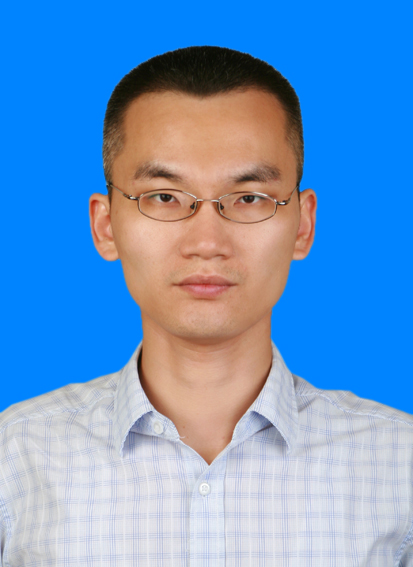}}]
{Ronghui Zhang} received a B.Sc. (Eng.) from the Department of Automation Science and Electrical Engineering, Hebei University, Baoding, China, in 2003, an M.S. degree in Vehicle Application Engineering from Jilin University, Changchun, China, in 2006, and a Ph.D. (Eng.) in Mechanical and Electrical Engineering from Changchun Institute of Optics, Fine Mechanics and Physics, the Chinese Academy of Sciences, Changchun, China, in 2009. After finishing his post-doctoral research work at INRIA, Paris, France, in February 2011, he is currently an Associate Professor with Guangdong Key Laboratory of Intelligent Transportation System, School of intelligent systems engineering, Sun Yat-sen University, Guangzhou, Guangdong 510275, P.R.China. His current research interests include computer vision, intelligent control and ITS.
\end{IEEEbiography}

\vspace{-9 mm}
\begin{IEEEbiography}[{\includegraphics[width=1in,height=1.25in,clip,keepaspectratio]{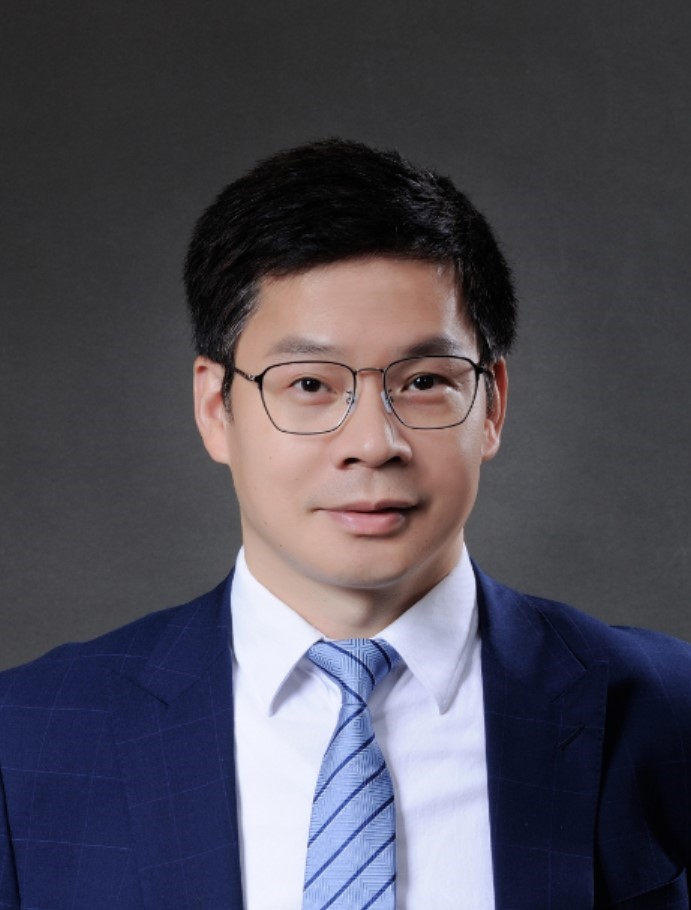}}]
{Nengchao Lyu} is a professor of Intelligent Transportation Systems Research Center, Wuhan University of Technology, China. He visited the University of Wisconsin-Madison as a visiting scholar in 2008. His research interests include advanced driver assistance system (ADAS) and intelligent vehicle (IV), traffic safety operation management, and traffic safety evaluation. He has hosted 4 National Nature Science Funds related to driving behavior and traffic safety; he has finished several basic research projects sponsored by the National Science and Technology Support Plan, Ministry of Transportation, etc. He has practical experience in safety evaluation, hosted over 10 highway safety evaluation projects. During his research career, he published more than 80 papers. He has won 4 technical invention awards of Hubei Province, Chinese Intelligent Transportation Association and Chinese Artificial Intelligence Institute.
\end{IEEEbiography}

\vspace{-9 mm}
\begin{IEEEbiography}[{\includegraphics[width=1in,height=1.25in,clip,keepaspectratio]{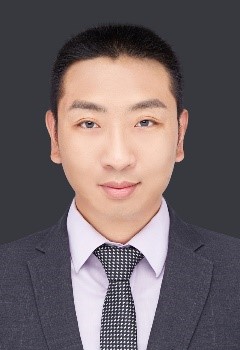}}]
{Yanyong Guo} received the M.S. degree in transportation engineering from Chang’an University, Xi’an, China, in 2012, and the Ph.D. degree in transportation engineering from Southeast University, Nanjing, China, in 2016. From 2014 to 2015, he was a Visiting Ph.D. Student with The University of British Columbia. He is currently a Professor with the School of Transportation, Southeast University. His research interests include road safety evaluations, traffic conflicts techniques, advanced statistical techniques for safety evaluation, and pedestrian and cyclist behaviors. He received the China National Scholarship, in 2014, and the Best Doctoral Dissertation Award from the China Intelligent Transportation Systems Association, in 2017. 
\end{IEEEbiography}

\vspace{-9 mm}
\begin{IEEEbiography}[{\includegraphics[width=1in,height=1.25in,clip,keepaspectratio]{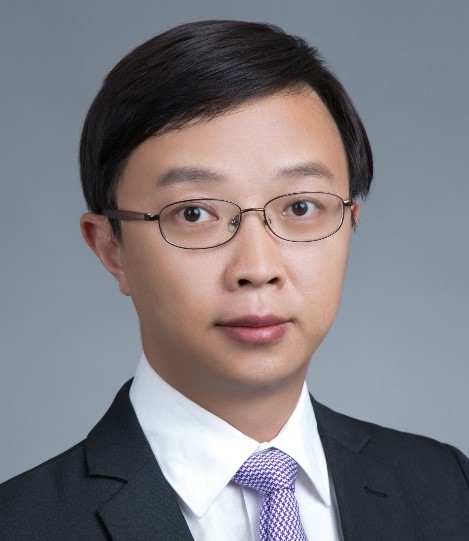}}]
{Hong-Ning Dai} is currently with the Department of Computer Science at Hong Kong Baptist University, Hong Kong as an associate professor. He obtained the Ph.D. degree in Computer Science and Engineering from Department of Computer Science and Engineering at the Chinese University of Hong Kong. His current research interests include the Internet of Things, big data, and blockchain technology. He has published more than 250 papers in top-tier journals and conferences with 19000+ citations. He has served as an associate editor for IEEE Communications Survey and Tutorials, IEEE Transactions on Intelligent Transportation Systems, IEEE Transactions on Industrial Informatics, IEEE Transactions on Industrial Cyber-Physical Systems, Ad Hoc Networks, and Connection Science. He is also a senior member of Association for Computing Machinery (ACM).
\end{IEEEbiography}

\vspace{-9 mm}
\begin{IEEEbiography}[{\includegraphics[width=1in,height=1.25in,clip,keepaspectratio]{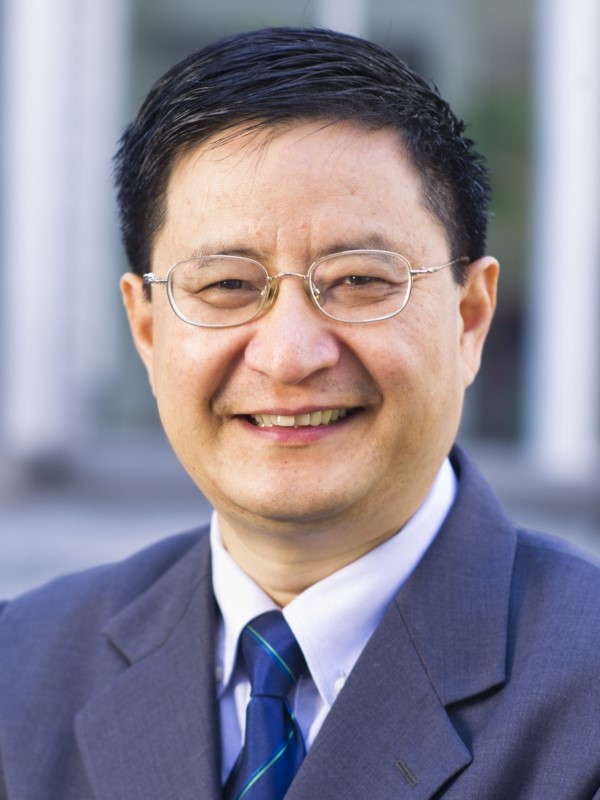}}]
{Hong Yan} received his PhD degree from Yale University. He was Professor of Imaging Science at the University of Sydney and currently is Wong Chun Hong Professor of Data Engineering and Chair Professor of Computer Engineering at City University of Hong Kong. Professor Yan's research interests include image processing, pattern recognition, and bioinformatics. He has over 600 journal and conference publications in these areas. Professor Yan is an IEEE Fellow and IAPR Fellow. He received the 2016 Norbert Wiener Award from the IEEE SMC Society for contributions to image and biomolecular pattern recognition techniques. He is a member of the European Academy of Sciences and Arts and a Fellow of the US National Academy of Inventors.
\end{IEEEbiography}

\vfill

\end{document}